\pdfoutput=1
\documentclass[12pt]{article}

\usepackage{graphicx}
\usepackage{subcaption}
\usepackage{listings}
\usepackage{amssymb}
\usepackage{amsmath}
\usepackage{bm}
\usepackage{cite}
\usepackage{url}
\usepackage{siunitx} 
\usepackage{tikz}
\usepackage[english]{babel}
\usepackage{algorithm}
\usepackage[noend]{algpseudocode}
\usepackage{acro}
\usepackage{setspace}
\usepackage[symbol, norule, flushmargin]{footmisc} 

\usepackage{geometry} 
\usepackage{secdot}
\sectiondot{subsection}

\usepackage{authblk}

\DeclareAcronym{deeprl}{short = deep-RL, long = deep reinforcement learning}
\DeclareAcronym{dqn}{short = DQN, long = deep Q-Network}
\DeclareAcronym{ddpg}{short = DDPG, long = deep deterministic policy gradient}
\DeclareAcronym{rl}{short = RL, long = reinforcement learning}
\DeclareAcronym{fov}{short = FoV, long = fields of view}
\DeclareAcronym{nn}{short = NN, long = neural networks}

 \title{\textbf{TOWARDS CONTINUOUS CONTROL OF FLIPPERS FOR A MULTI-TERRAIN ROBOT USING DEEP REINFORCEMENT LEARNING}}

\author[*]{Giuseppe Paolo}
\author[** $\dagger$]{Lei Tai}
\author[**]{Ming Liu}
\affil[*]{Department of Mechanical and Process Engineering, ETH Z{\"u}rich.}
\affil[**]{Department of ECE, The Hong Kong University of Science and Technology.}
\affil[$\dagger$]{Department of MBE, City University of Hong Kong.}

\date{}

\begin{document}
\pagenumbering{gobble}
\thispagestyle{empty}
\maketitle

\section*{Abstract}
In this paper we focus on developing a control
algorithm for multi-terrain tracked robots with flippers using a
\ac{rl} approach. The work is based on the \ac{ddpg} algorithm, proven to be
very successful in simple simulation environments. The algorithm
works in an end-to-end fashion in order to control the continuous
position of the flippers. This end-to-end approach makes it easy
to apply the controller to a wide array of circumstances, but the
huge flexibility comes to the cost of an increased difficulty of
solution. The complexity of the task is enlarged even more by the fact that real multi-terrain robots move in partially observable environments. Notwithstanding these complications, being able to smoothly control a multi-terrain robot can produce huge benefits in impaired
people daily lives or in search and rescue situations.

\footnotetext{Manuscript received 22 September 2017}
\clearpage
\section*{Key Words}
Deep reinforcement learning, multi-terrain robot, deep deterministic policy gradient, neural networks, end-to-end learning

\pagenumbering{arabic}

\section{Introduction}
Tracked multi-terrain robots are a typology of semi-autonomous robots that are getting more and more important thanks to their flexibility. They can be valuable in a wide range of situations, from wheelchairs\footnote{http://b-free.hk/2017/} to search and rescue applications. This is due to the fact that they can overcome obstacles of any kind quite easily, but at the same time be more stable, and thus safer, than their legged counterparts.

In order to give greater capabilities to these vehicles additional components, like flippers, need to be used, thus making the locomotion system more complex and the control of the robot harder. This increased complexity needs to be taken care of in order to improve the user's experience. From here the need for more powerful and flexible control software. Its development is the focus of this work.

Using state of the art \ac{deeprl} algorithms, we developed an end-to-end controller, capable of reading raw data from the camera and calculate the best positions for the flippers, in order to tackle the obstacles found along the robot's path. This allows fast responses while at the same time providing greater stability to the platform.
The use of \ac{deeprl} is promising for this task because it stimulates the robot, through the optimization of a reward function, to find the best way of action in order to achieve its goal.
If this reward function is well designed, it can lead to better performances with respect to humans. Additionally, thanks to the algorithm's ability at generalization, it can learn to solve many different problems, as shown in \cite{Mnih2015}. This means that it can be applied, without excessive modifications, to different settings and environments.

To train the algorithm, we simulated the robot in a virtual environment, letting it learn how to use the flippers to climb a flight of stairs. It is not easy to find solutions in situations like these, where the environment is intrinsically partially observable, through \ac{deeprl}. At the same time, though, these settings are very similar to the real world scenarios where the vehicle will be operated in. In fact, an important limiting factor for the application of \ac{deeprl} algorithms to the real world is given by the fact that fully observable environments, in which \ac{deeprl} performs very well, are seldom found in everyday life. Notwithstanding these issues, their application can open new and exciting opportunities for robotics, while making many people's lives easier. 	

The work is structured so that section \ref{sec:work} describes some related works; then section \ref{sec:robot} shows the robot structure and the sensors it is equipped with; \ref{sec:dl} presents the algorithm; section \ref{sec:impl} describes the implementation details; section \ref{sec:res} presents the results; finally, section \ref{sec:disc} discusses the implications of our approach while conclusions and future improvements are discussed in \ref{sec:con}. Finally we provide additional reference to related material.

\section{Related Work}\label{sec:work}
The state of the art \ac{deeprl} algorithms manage to solve many different tasks. Among them, the two most widely used are \ac{dqn}\cite{Mnih2015} and \ac{ddpg}\cite{ddpg}. The former managed to master, often better than human players, many Atari games just using raw pixel images as input and outputting discrete actions. The latter instead achieved something more complex: the use of continuous actions. This could be the key to extend \ac{deeprl} to real environments: in these cases in fact, just using discrete actions might not be enough to guarantee good performances. Another important result was obtained in \cite{tensegrity}, where the authors applied \ac{deeprl} to a real and complex robot to make it learn how to move in the most efficient way. It needs to be noticed though among these works, only the first two used \ac{deeprl} in an end-to-end fashion only in game-style simulators with mostly fully observable environments.

A problem of \ac{rl} is the time required for the training and for the collection of a sufficient number of samples. This issue has been addressed in \cite{a3c}, where agents working in parallel have been used for sample collection, thus significantly speeding up the network's training. 
In \cite{tai2017virtual} the authors applied an asynchronous \ac{deeprl} algorithm to a mobile robot to navigate it through unseen environments, obtaining better results than previously available algorithms. 

When solving complex problems with \ac{rl}, having the algorithm converge can be tricky. To solve this issue many tricks have been developed, like the use of Replay Memory\cite{replay} and auxiliary target networks\cite{Mnih2015}. Another trick used in \cite{auxiliary} is to give the agent other auxiliary tasks, which differ from the main goal, to solve; this pushes the agent to explore more, thus leading to better solutions.

If \ac{rl} algorithms are to be applied to real-world robots, the risk that they can cause damage to the environment and the robot itself while exploring different policies, needs to be considered. To overcome this problem Pecka et al.\cite{constr} and Zimmermann et al.\cite{constr2} added some constraints to the exploration policies. Although they also used a tracked robot with flippers, this solution is not feasible when using \ac{nn} because, as of now, it is not possible to control their learning process through constraints. Thus, in our work we relied on training the agent in a simulation environment. This allowed the algorithm to test all the policies needed with no real harm.

\section{Preliminaries} \label{sec:background}
In this section, we present the robot we used, the simulation environment and the algorithm itself.
\subsection{Robot} \label{sec:robot}
\begin{figure}
	\centering
	\begin{subfigure}[b]{85mm}
		\includegraphics[width=85mm]{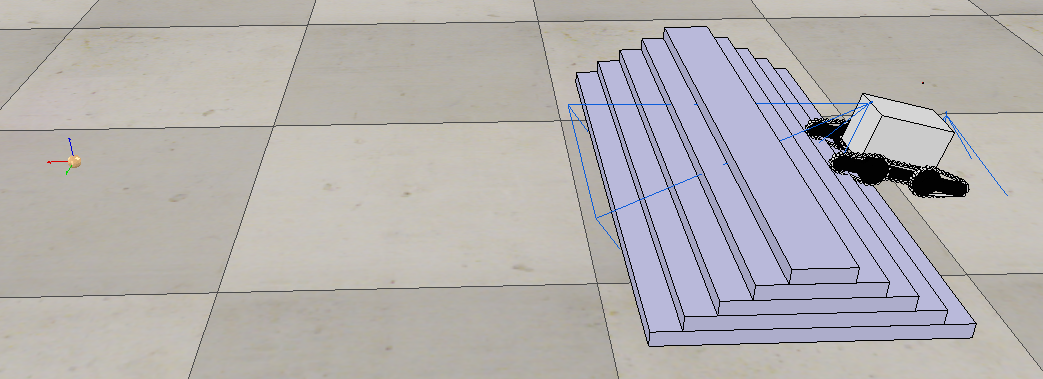}
		\caption{}
		\label{fig:robot_sim} 
	\end{subfigure}
	
	\begin{subfigure}[b]{85mm}
		\includegraphics[width=85mm]{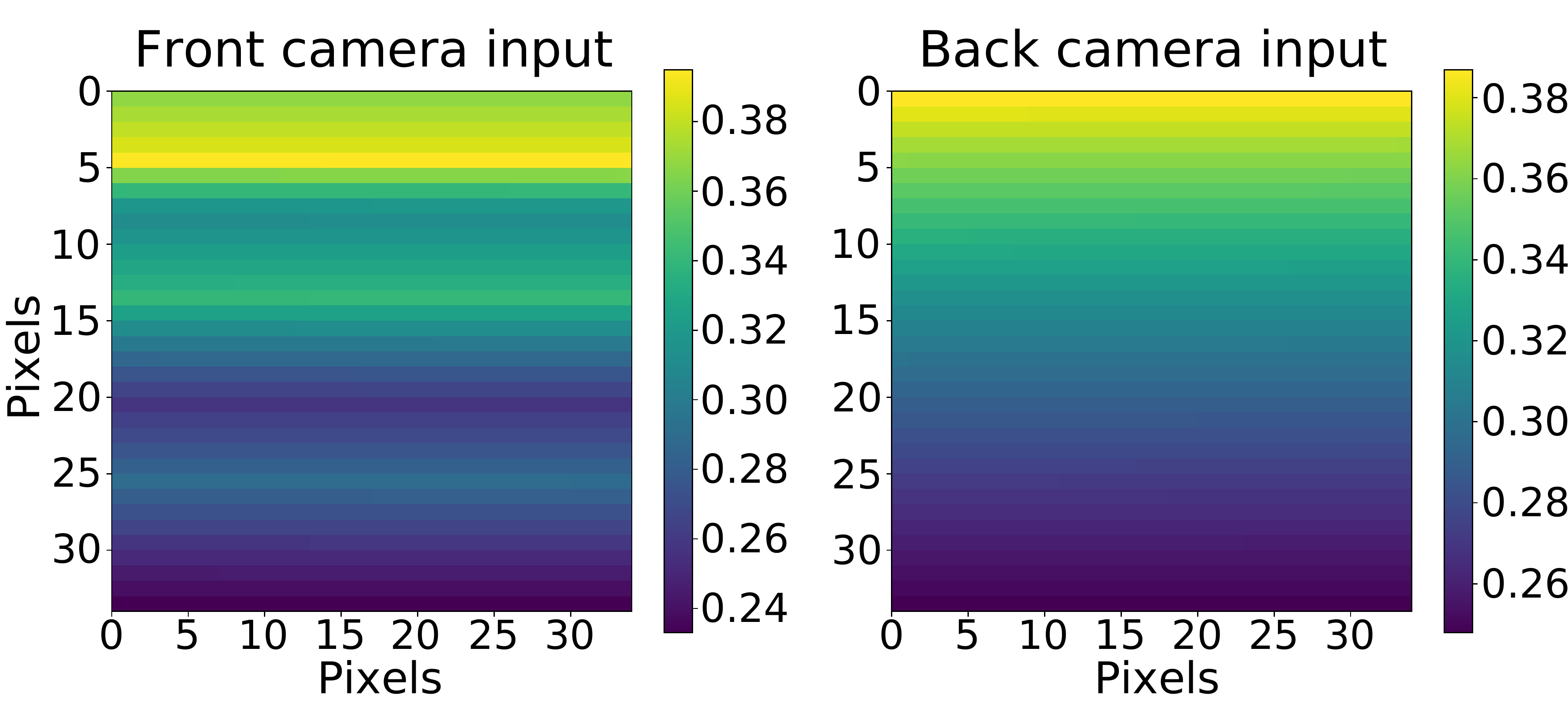}
		\caption{}
		\label{fig:cameras}
	\end{subfigure}
	
	\caption[Two numerical solutions]{(a) \textit{Simulation environment. On the left is possible to see the goal point.} (b) \textit{Input from the two depth cameras while the robot is trying to climb the stairs as in (a).} }
\end{figure}

The robot is equipped with tracks and flippers, two in the front and two in the rear. During our work, we created a simulated version of it in order to be able to safely train the algorithm. The robot has been simulated in V-Rep\footnote{http://www.coppeliarobotics.com/} and is based on the tracked wheelchair by B-Free\footnote{http://b-free.hk/2017/}. In Fig. \ref{fig:robot_sim} it is possible to see the robot model, with the four flippers and the cameras, in the simulation environment while it is trying to climb a flight of stairs. As shown in the figure, the robot is equipped with two depth cameras, one in the front and one in the back, whose \ac{fov} volumes are delimited by the blue lines and are of the same dimensions of a Kinect\footnote{https://en.wikipedia.org/wiki/Kinect} \ac{fov}. Cameras' resolution is $34 \times 34$ pixels.

In addition to cameras, the sensing equipment is completed by a 6 degree IMU. This sensor is mounted on the central vertical axis of the platform and is composed of a 3D gyroscope and a 3D accelerometer: this allows the robot to know its own inclination and acceleration at any given moment during the operations.

 Fig. \ref{fig:cameras} shows the inputs coming from the two vision sensors while the robot is in the configuration shown in Fig. \ref{fig:robot_sim}.  On the left, there are the data coming from the front camera. It is possible to see the differences in depths given by the stairs. On the right, are shown the back camera's readings, focusing on the ground, thus showing just a smooth changing in depth.

The four flippers can change their orientation between $+90\si{\degree}$ and $-90\si{\degree}$ from the horizontal position, considered as $0\si{\degree}$. It is possible to control each flipper separately, but in order to reduce the action space, from four dimensions to two, and also to make the simulation more similar to the real platform, right and left flippers were paired. 

\subsection{Deep Learning} \label{sec:dl}
In order to address problems with \ac{rl}, we need to look for policies and states values. These values need to be stored somewhere. There are two ways to do it: in a hash table or in a function approximator. The former works adequately for trivial problems, being easier to implement and allowing to know at every moment the values associated with each state. However, in more complex situations, with high-dimensional observations and actions, the table can become quite resource intensive\cite{survey}\cite{sutton1998reinforcement}. From here the need to use more elaborate but also more powerful function approximators to store and retrieve these values. 
In deep learning, \ac{nn} are used as function approximators, allowing the handling of continuous actions, as done in \ac{ddpg}. The aforementioned reason is why our project is based on this algorithm in order to develop a good controller for our robot. 

\ac{ddpg} uses two networks: the Critic net, $\theta^{Q}$, for the Q value, and the Actor net, $\theta^{\mu}$, for the actions. To help the \ac{nn}s' convergence other two networks are used, called target networks, whose outputs are used as targets for training the Critic and Actor ones. The weights of these two target networks are updated with soft updates according to:
\begin{equation}
\theta' = \tau\theta + (1 - \tau)\theta'  
\end{equation}
where $\theta'$ are the parameters of the target nets and $\theta$ the ones of the two main \ac{nn}. The weights of the main \ac{nn} are updated through stochastic gradient descent.

This algorithm, in a fully observable environment, is capable of obtaining good policies in an end-to-end fashion. That means it can just be fed with raw data, without pre-elaboration and, thanks to the capabilities of \ac{nn}s, obtain as outputs appropriate actions for each situation. What we want to achieve is to find a way to apply this power to more complex, real-world-like situations. The flow of \ac{ddpg} is shown in Algorithm \ref{alg}.

\begin{algorithm}
\caption{\ac{ddpg}}\label{alg}
\begin{algorithmic}

\State {Randomly initialize Critic net $\theta^{Q}$ and Actor net $\theta^{\mu}$} 

\State {Initialize target nets: $\theta^{Q'} = \theta^{Q}$ and $\theta^{\mu'} = \theta^{\mu}$}

\State {Initialize replay buffer $R$}

\For{ episode = 1, M}	
	\For{ step = 1, T}
		\State {Do action}
		\State {Observe state}
		\State {Save state in $R$}
		\State {Sample minibatch of size $m$ from $R$}
		\State {Update networks}
		\State {Obtain next action from nets}
		\If {GoalReached \textbf{or} Failed}
			\State {End episode}
		\Else { Continue}
		\EndIf		
	\EndFor
\EndFor
\end{algorithmic}
\end{algorithm}

\section{Implementation}\label{sec:impl}
In this section, we formalize, from a mathematical point of view, the problem we want to solve and describe in detail how we approach its solution.
\subsection{Problem Definition}
In this paper we aimed at implementing an end-to-end algorithm capable of reading raw sensor data and, after some elaboration, help to control a tracked robot. So what we are looking for is a function in the form:

\begin{equation}
\textbf{a}_{t} = f(\textbf{s}_{t}, \textbf{a}_{t-1})
\end{equation}

where $\textbf{a}_{t} = \{a_{t}^{front}, a_{t}^{rear}\}$ is the vector of actions, with ${a_{t}^{i} \in [-1,1]}$ being the action taken at time $t$ and $\textbf{s}_{t}$ the state the robot is in at time $t$. 

The state is defined as:
\begin{equation}
\textbf{s}_{t} = [ c^{front}_{t}, c^{back}_{t}, w_{t} ] 
\end{equation}

with $(c^{front}_{t}, c^{back}_{t})$ being the reading at time $t$ of the front and back depth cameras respectively, and $w_{t}$ being the IMU data at time $t$. 

The data coming from the cameras are shaped as a 3-dimensional matrix containing four frames captured at $0.25s$ time intervals. This way the robot can infer its velocity and the changes in its orientation and inclination due to the actions it is performing. A similar strategy is also used in \cite{Mnih2015} and \cite{ddpg}. At the same time, the data coming from the IMU is a 6-dimensional vector: each element of this vector is calculated as the average of four values captured at the same time instant as the cameras' frames. Using this running average over the IMU data we manage to smooth out the intrinsic noise of the sensor while still retaining the important informations.
This is the only pre-elaboration done on the data.

\subsection{Network Structure}
The networks used are Multi-Input Single-Output nets: the Critic is composed of two convolutional branches, one for each camera, and two fully connected branches for the IMU data and the previous actions respectively. All these branches are then merged together and passed to three fully connected layers that will output the Q value for the state-action input pair. The Actor net structure is the same as the Critic with the exception that there is no previous actions input. The output of this network is the 2-dimensional action vector where the first element controls the position of the front flippers and the second one drives the rear flippers. 

\begin{figure}[h]
	\centering
	\includegraphics[width=80mm]{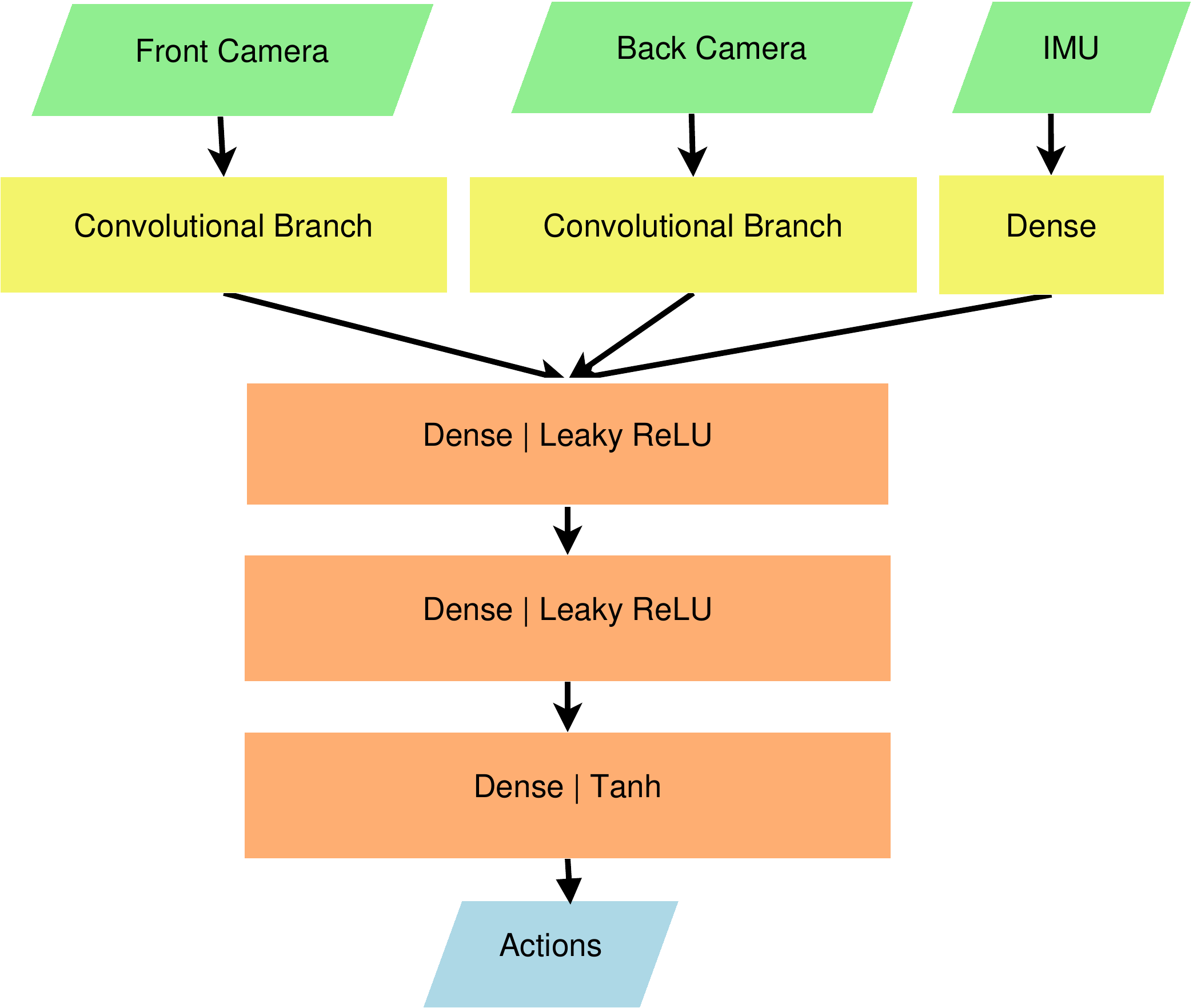}
	\caption{\textit{Actor network model. The three inputs are shown in light green, while their relative branches are in yellow. The fully connected layers are in orange and, finally, the output is represented in light blue.}}
	\label{fig:actor}
\end{figure}

\begin{figure}[h]
	\centering
	\includegraphics[width=80mm]{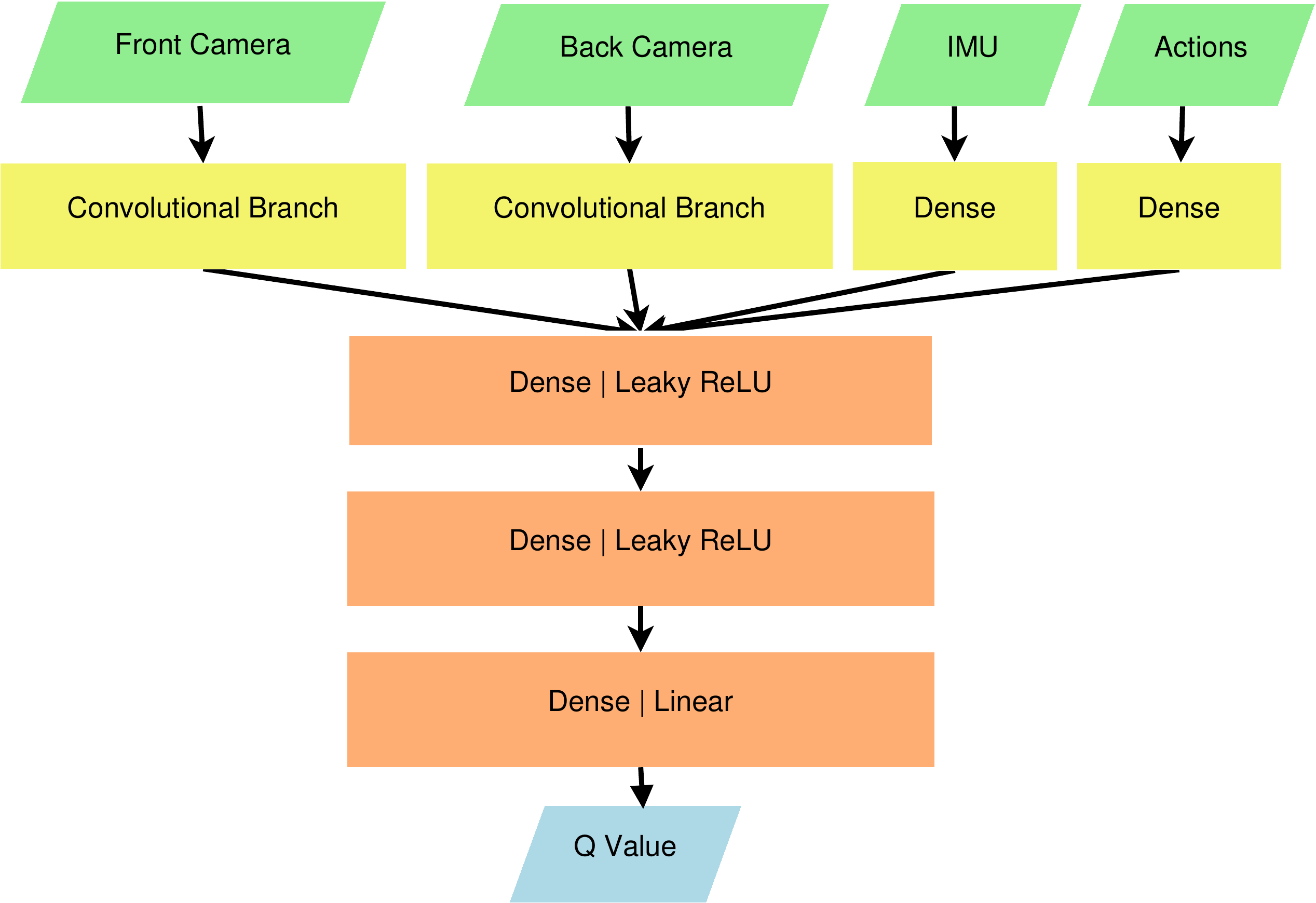}
	\caption{\textit{Critic network model. The four inputs are shown in light green, while their relative branches are in yellow. The fully connected layers are in orange and, finally, the output is represented in light blue.}}
	\label{fig:critic}
\end{figure}

The activations of the convolutional branches are linear. The fully connected layers are instead equipped with Leaky-ReLU activation functions, exception made for the last layer.
This one, in fact, is different among the two networks: the Critic has a linear output activation, while the Actor output goes through a $tanh$ function. This difference is due to the fact that we want to limit the actions in the range $[-1,1]$ in order to be able to control the flippers, while the Q value can be any value in $\Re$.

A representation of the two networks is shown in Fig. \ref{fig:actor} and \ref{fig:critic}, where the inputs are represented in light green and the outputs in light blue.

\subsection{Reward Function}
The reward function is the most important part of a \ac{rl} algorithm, defining the goal of the training process. A badly designed one can hinder learning and result in totally wrong policies. This means that conceiving a proper reward function is not an easy task, but requires a lot of effort. Nevertheless, it remains easier to create compared to an efficient controller and is more prone to generalization. This is the main reason why reinforcement learning is becoming increasingly more popular. In this paper, we base our reward on the data from the IMU and the distance traversed by the platform.

First, we take the square of the gyroscope data, that gives the velocity of rotation around each axis, and penalize it, in order to prevent excessive swinging that could cause harm of any sort to the platform during operation.
Then, in the same way, we use the data coming from the accelerometer, that is the acceleration along each axis, to penalize the inclination with respect to the horizontal plane. 
This can be expressed as:

\begin{equation} r_{t}(s_{t}) = -\theta_{g_{x}}g_{x}^{2} - \theta_{g_{y}}g_{y}^{2} - \theta_{g_{z}}g_{z}^{2} - \theta_{a_{x}}a_{x} - \theta_{a_{y}}a_{y} 
\label{eq:r}
\end{equation}

where $\{g_{x}, g_{y}, g_{z} \}$ represent the values of the gyroscope along the 3 axes $\{x, y, z\}$, $\{a_{x}, a_{y}, a_{z} \}$ the values of the accelerometer, and  the vector $\bm{\theta} = \{\theta_{g_{x}}, \theta_{g_{y}}, \theta_{g_{z}}, \theta_{a_{x}}, \theta_{a_{y}}\}$ contains the parameters used to scale the data according to their importance to us.

Another situation we penalize heavily, for obvious reasons, is the case in which the robot flips upside-down. In this circumstance, in fact, we give it a $-1000$ penalty and put an end to the current episode. At the same time, we give a reward proportional to how much the platform advanced when moving forward and slightly punish it, with $c_{stuck} = 5$, for every time step it gets stuck. Finally, when it reaches the goal point it gets a reward of $+100$. 

The resulting function can be expressed in the following way:
\begin{equation} r_{t}^{final}(s_{t}) = 
	\begin{cases}
	r_{t}(s_{t}) + c_{mov}\delta_{t},& \text{if } \delta_{t}>0\\
	r_{t}(s_{t}) - c_{stuck}, & \text{if } \delta_{t}=0\\
	-1000, & \text{if } c_{flip} == True\\
	+100, & \text{if } c_{goal} == True
	\end{cases}
\end{equation}

where $\delta_{t} = d_{t} - d_{t-1}$, with $d_{t}$ being the travelled distance until time $t$. $c_{flip}$ and $c_{goal}$ are boolean flags: the former is set to $True$ if the robot flips over while the latter is set to $True$ if the robot manages to climb the stairs and reach the goal point. $c_{mov}$ is the reward scaling factor for the forward movement distance.

\subsection{Learning Setup}
As all machine learning algorithms, also our implementation has many hyperparameters to set besides the networks' structure. In our experiments, we used the Adam \cite{adam} optimizer with a learning rate of $10^{-6}$ for both networks. The training was done in a synchronous online fashion with the data collected during the training process itself and stored into the replay memory. This memory had a capacity of 45000 elements composed of the observation from 4 timesteps. 

In order to push the algorithm to explore the action space, some noise sampled from an Ornstein-Uhlenbek process was added to the actions. 
The target networks update rate was set to $10^{-5}$. Moreover, the maximum length of each episode was constrained to 150 steps.

\section{Results}\label{sec:res}
\begin{figure}[h]
	\centering
	\includegraphics[width=80mm]{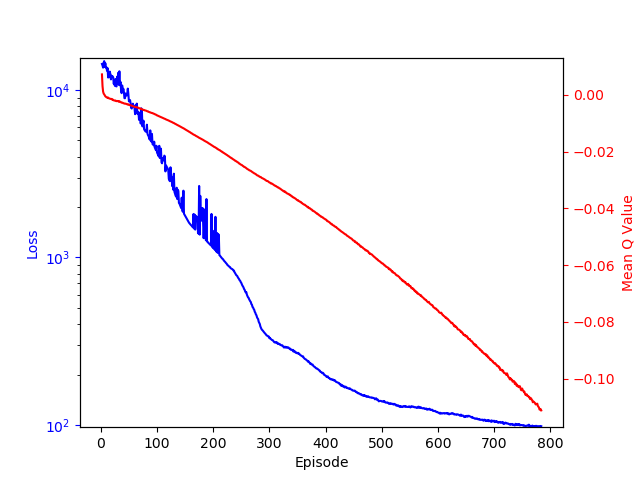}
	\caption{\textit{2D convolutional neural network. The loss is expressed in logarithmic scale in order to represent it better, while for the Q value a linear scale was used.}}
	\label{fig:2dcnn}
\end{figure}

\begin{figure}[h]
	\centering
	\includegraphics[width=80mm]{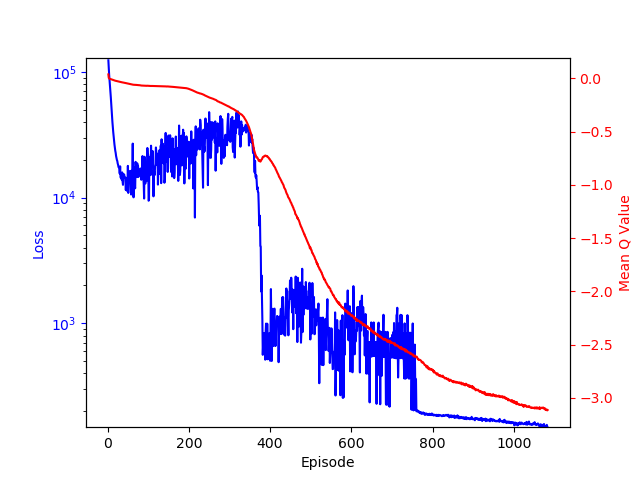}
	\caption{\textit{3D convolutional neural network. The loss is expressed in logarithmic scale in order to represent it better, while for the Q value a linear scale was used.}}
	\label{fig:3dcnn}
\end{figure}

In order to understand which topology of neural networks was the most appropriate, different configurations were tested and compared: 3D convolutional nets, that let the time dimension intact till the final dense layers, and standard 2D convolutional ones. The training procedure was done in the V-Rep simulation environment. The results are shown in Fig. \ref{fig:3dcnn} and \ref{fig:2dcnn}.

As is possible to see, the 2D convolutional networks perform much better than their 3D counterpart, in fact, the loss for the first topology decreases faster and in a smoother fashion. At the same time, the second network suffers from a huge decrease in the Q value, indicating its inability to optimize the policy.

\section{Discussion}\label{sec:disc}
In this work, we started developing a framework for applying deep reinforcement learning to real robots. As already mentioned, this is a hard problem to solve due to the partial observability of the real world. In fact, previous attempts to use deep-RL focused on fully observable environments, where it is easier for the algorithm to obtain good results, but real-world environments are intrinsically partially observable. 

\begin{figure}[h]
	\centering
	\includegraphics[width=85mm]{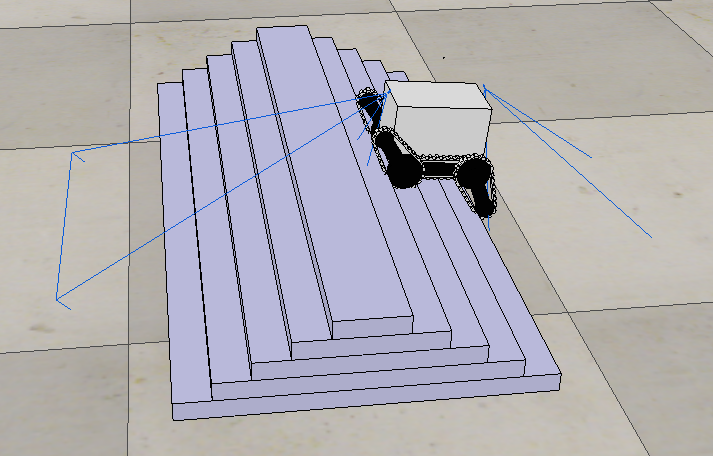}
	\caption{\textit{Robot stuck in local minima.}}
	\label{fig:stuck}
\end{figure}

During our experiments, we found out that the robot tended to get stuck in a local minimum with the front flippers at $+90\si{\degree}$ and the back one at $-90\si{\degree}$. This situation can be seen in Fig. \ref{fig:stuck}.

This configuration is quite effective to climb the first few steps of a flight of stairs but results highly inefficient in order to finish the task. From here it can be seen that DDPG did not prove to be efficient for the task at hand. This can be given by many factors that will be analyzed in the following parts, while suggesting strategies to improve the results.
\subsection{Data Representation and Networks Structure}
One of the causes at the root of the bad policy case could be the data representation. In fact, the algorithm has to learn the actions to take from a sequence of previous steps. This task requires the networks to have some kind of memory structure that could work with the time dimension of the sequence. In our case, this memory is faked through the use of 4 following frames for each state, from which the nets should infer velocity and movements. This means that the time frames are elaborated by feed-forward networks, that are not perfectly fit for the task. 
In fact, as can be seen from Fig. \ref{fig:2dcnn} and \ref{fig:3dcnn}, the loss decreases, implying the algorithm is learning something but it is not optimal, as can be noted from the decreasing Q value and from visual inspection of the simulation.

Moreover, given the higher performance of the 2D network, compared to the 3D one, we can infer that a proper time dimension manipulation is a fundamental requirement of the process. In fact, in the former topology, this dimension is elaborated by the convolution branches, while in the latter this elaboration is delayed till the fully connected layers.

Thus the application of \ac{nn} specifically designed to handle time series might significantly improve the results of the policy search. This kind of networks are called recurrent \ac{nn} and the most promising of this kind are LSTM\cite{lstm}. Using recurrent nets would eliminate the need for the 4 frames while at the same time allowing to work with whole training episodes, thus removing the need for the memory replay.

\subsection{Reward Function}
As stated earlier, the most important factor in \ac{rl} is the reward function. This function is what tells the algorithm what it has to learn and how, how to interpret the data and if it is acting properly or not. This is why it requires a lot of study and experimentation to find the right one and have a properly converging algorithm. 

Our reward function describes what we thought to be the most important aspects to take in consideration in order to solve the problem. to guarantee the safety and well-being of the user and of the platform itself we penalized excessive swinging during operation through the definition of \eqref{eq:r}. In this equation, we take into account the data coming from the IMU mounted at the centre of the platform.

Other than that, situations in which the flippers are in a position that does not allow the platform to proceed are also penalized. This penalty is very important toward the accomplishment of the task, given that the platform is equipped with flippers in order to overcome this kind of situations.

Changing the reward function or improving it, taking into considerations also other factors like the distance from the stairs or the flippers' inclination, could lead to performance improvements and help in solving the task.

\section{Conclusions}\label{sec:con}
In this work, we have set the basis for applying \ac{deeprl} algorithms to interesting real-world problems. This task is very complex given the partial observability of the environment and also given the high processing power such kind of algorithms require in order to learn good policies. Nonetheless, a proper implementation of these techniques could improve the lives of many people around the world.

Future work includes testing new cost functions and switch from feed-forward neural networks to recurrent ones, namely LSTM, thus getting rid of the multi-frames inputs. Other interesting research directions could be to discard DDPG in favor of continuous DQN, a different algorithm, or to totally switch from continuous actions to discrete ones, where the action network output is a flag that controls if we have to increment or decrement the flippers' angles of a fixed quantity. If this quantity is small enough the movements will be so smooth they will seem like a continuous movement. Moreover, this approach would simplify a lot the action space, making the task easier to solve.

Another line of work could be to use imitation \ac{rl}, where the algorithm is trained by a human teacher showing it how to perform the preferred actions. Once the algorithm has learned them, then it could keep optimizing from that starting point. This should lead to improved and super-human performances.

\nocite{*}

\bibliographystyle{ieeetr}

\addcontentsline{toc}{chapter}{Bibliography}
\cleardoublepage

\end{document}